%% file: root.tex
\documentclass[letterpaper, 10 pt, conference]{ieeeconf}  % Comment this line out if you need a4paper

\IEEEoverridecommandlockouts                              % This command is only needed if 
\usepackage{graphics} %
\usepackage{epsfig} %
\usepackage{amssymb}  %
\usepackage{bm}
\usepackage{algpseudocode}
\usepackage{algorithm}
\usepackage{xcolor}
\usepackage{amsmath}
\usepackage{lineno}
\makeatletter
\let\NAT@parse\undefined
\makeatother
\usepackage[colorlinks=true, citecolor=green, linkcolor=red]{hyperref} 
\usepackage{multirow}
\usepackage{booktabs}
\usepackage{cite}
\usepackage{caption}
\usepackage{subcaption}
\usepackage{bm}
\usepackage{bbm}

\title{\LARGE \bf
RoboReflect: A Robotic Reflective Reasoning Framework for Grasping Ambiguous-Condition Objects
}

\author{Zhen Luo$^{1,3 \dag}$, Yixuan Yang$^{1,4 \dag}$, Yanfu Zhang$^{5}$ and Feng Zheng$^{1,2*}$% <-this % stops a space
\thanks{\dag Equal Contribution. *Corresponding author.}% <-this % stops a space
\thanks{$^{1}$Department of Computer Science and Engineering, SUStech, China
        }%
\thanks{$^{2}$Institute of Multiple Agents and Embodied Intelligence, Peng Cheng Laboratory, China}
\thanks{$^{3}$Shanghai Innovation Institute, China
        }%
\thanks{$^{4}$Department of Computer Science and Engineering, University of Warwick, Coventry, UK.}
\thanks{$^{5}$Department of Computer Science, College of William and Mary, USA.
        }%
}

\begin{document}

\maketitle
\thispagestyle{empty}
\pagestyle{empty}
%\vspace{-2em}

%%%%%%%%%%%%%%%%%%%%%%%%%%%%%%%%%%%%%%%%%%%%%%%%%%%%%%%%%%%%%%%%%%%%%%%%%%%%%%%%
\begin{abstract}
As robotic technology rapidly develops, robots are being employed in an increasing number of fields. 
However, due to the complexity of deployment environments or the prevalence of ambiguous-condition objects, the practical application of robotics still faces many challenges, leading to frequent errors. 
Traditional methods and some LLM-based approaches, although improved, still require substantial human intervention and struggle with autonomous error correction in complex scenarios.
In this work, we propose RoboReflect, a novel framework leveraging large vision-language models (LVLMs) to enable self-reflection and autonomous error correction in robotic grasping tasks.  
RoboReflect allows robots to automatically adjust their strategies based on unsuccessful attempts until successful execution is achieved.
The corrected strategies are saved in the memory for future task reference.
We evaluate RoboReflect through extensive testing on eight common objects prone to ambiguous conditions of three categories.
Our results demonstrate that RoboReflect not only outperforms existing grasp pose estimation methods like AnyGrasp and high-level action planning techniques ReKep with GPT-4V but also significantly enhances the robot's capability to adapt and correct errors independently. 
These findings underscore the critical importance of autonomous self-reflection in robotic systems while effectively addressing the challenges posed by ambiguous-condition environments.

\end{abstract}

%%%%%%%%%%%%%%%%%%%%%%%%%%%%%%%%%%%%%%%%%%%%%%%%%%%%%%%%%%%%%%%%%%%%%%%%%%%%%%%%
\input{sections/introduction}

\input{sections/related_work}
\input{sections/methodology}

\input{sections/experiments}

\bibliographystyle{IEEEtran}
\bibliography{IEEEabrv}

\end{document}

%% file: sections/introduction.tex
\section{INTRODUCTION}

With the rapid development of artificial intelligence, embodied AI and robotics technology have made significant strides across various fields. A notable instance is robotic arm grasping tasks, which have become a vital application in industrial automation and service robotics~\cite{wang2024llm,guan2024atom,macaluso2024toward}. The recent rise of Large Language Models (LLMs) has further broadened the potential of these technologies. 
Compared to traditional methods, LLM-based robotic arms can understand tasks from natural language instructions and autonomously generate grasping policies. 
This capability makes robotic arms more flexible when facing diverse and dynamically changing tasks.
Furthermore, this also lays the foundation for robotic arms to deal with complex, long-horizon manipulation tasks. 
Therefore, researchers are now exploring ways to integrate LLMs into robotic systems, envisioning them as the ``brain" of embodied intelligence, capable of facilitating actions in the real world.

In real-world scenarios, the properties of daily necessities often differ from those in experimental training data (e.g., Empty tissue bags, deformed bottles, and fragile toys).
Some objects whose properties change with use undergo corresponding changes at different conditions, consequently resulting in errors in the grasping (\textit{e.g.}, fulfilled tissue bags and empty tissue bags).
Additionally, achieving reliable grasping performance necessitates deeper understanding of objects due to factors such as material composition or edibility (e.g., ice cream on a stick).

We refer to such objects as \textit{ambiguous-condition objects}, where both robotic arm systems and LLMs struggle to determine the object's current condition to calculate the affordance point or grasp pose at the first time, often leading to grasping failures or unreasonable actions. 
Therefore, whether applying traditional methods or LLM-driven approaches, distinguishing between objects with the same label but differing physical properties remains highly challenging for ensuring proper grasping.
How to reduce these misjudgments remains a key problem in more general and practical applications.

Some existing studies attempted to teach embodied robots with the capability to correct errors, making them more robust.
REFLECT~\cite{liu2023reflect} primarily addresses errors arising from the robot's planning and execution but struggles with understanding complex and ambiguous real-world scenarios.
YAY Robot~\cite{shi2024yell}, which incorporates language correction via human feedback, demonstrates robustness across diverse environments. 
However, it heavily relies on human intervention during the error correction process.
While these methods broaden the robot's adaptability, they fail to develop the robot's ability to autonomously identify and resolve ambiguous-condition objects.
Ultimately, these techniques rely on human guidance to enhance adaptability, rather than activating the robot's autonomous error-solving capabilities when encountering objects in ambiguous conditions.

To address the above challenges, we introduce a novel framework named RoboReflect, a first-of-its-kind tailored solution for embodied robot autonomous reflection and error correction using Large Vision-Language Models (LVLMs) to grasp objects under ambiguous conditions.
In this framework, LVLM provides high-level policies for grasping tasks to embodied robots based on instructions and the environment and decomposes them into executable primitive actions to guide the robot in completing the task.
After each action, LVLM checks whether it was successfully executed; if it fails, LVLM reflects on the reasons for failure based on the results of the failed attempt and corrects the action strategy until the task is successfully completed.
The main contribution of the RoboReflect can be summarized as follows:
\begin{itemize}

\item We introduce a novel self-reflect and correction robot grasping framework, RoboReflect, which can autonomously correct the failure of the grasping task on ambiguous-condition objects.

\item RoboReflect utilizes the Reflective Reasoning module with LVLM and a Memory module, enabling the framework to resolve the ambiguous grasping scenario and incorporates a memory mechanism to handle ambiguous-condition issues that have already been encountered.

\item We categorize ambiguous-condition objects into three types, and test RoboReflect on eight common objects. Ours consistently outperforms ReKep~\cite{huang2024rekep}, AnyGrasp~\cite{fang2023anygrasp} and GPT-4V~\cite{achiam2023gpt} driven grasp method, which highlights the critical role of autonomous reflection-correction in ambiguous grasping scenarios.
\end{itemize}

%% file: sections/related_work.tex
\section{RELATED WORK}
\noindent \textbf{Large Models for Robotics.}
Recent advancements in LVLMs like GPT-4V\cite{achiam2023gpt}, PaLM\cite{chowdhery2023palm}, and LLaMA\cite{touvron2023llama} have spurred interest in integrating these models into robotics. Some research focuses on generating high-level task plans using human language instructions to guide robots in executing complex tasks\cite{ahn2022can,lin2023text2motion}, while others explore the direct generation of robotic control code\cite{singh2023progprompt,liang2023code}. Additionally, LVLMs are fine-tuned on robotic datasets to learn and adapt actions in real-time environments\cite{kim2024openvla,zhen20243d,li2023vision}. While these efforts enhance robotic comprehension and planning, real-world task execution remains challenging. We propose leveraging LVLMs' inferential abilities for autonomous reflective reasoning to address these complexities.

\noindent \textbf{Robot Action Correction.}
Recent studies have explored using LLMs for error reasoning and action correction in robots\cite{shinn2303reflexion,liu2023reflect}, but most focus on simple action failures with limited environmental understanding. Other approaches rely on human feedback to correct robot actions\cite{huang2022inner,ren2023robots}, but these often require extensive human intervention to handle diverse errors in complex environments. To tackle this, we propose the RoboReflect framework, which enables robots to address errors autonomously through reflection and recommendation mechanisms, improving their adaptability and robustness in ambiguous scenarios.

%% file: sections/methodology.tex
\section{METHODOLOGY}

\begin{figure*}[tbp!]
    \centering
    \includegraphics[width=0.92\textwidth]{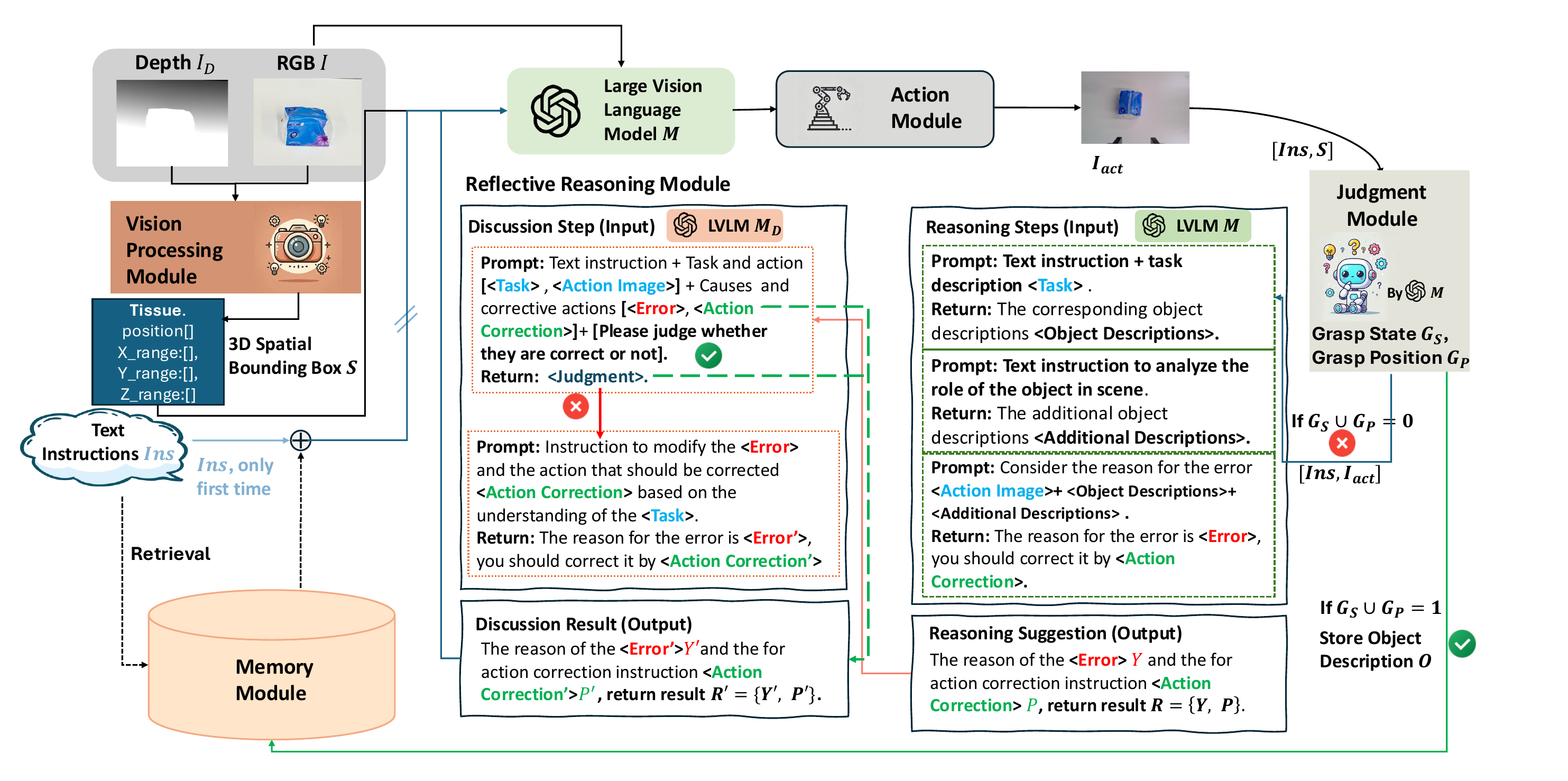}
    \vspace{-0.5em}
    \caption{\small{
\textbf{The RoboReflect Framework for Autonomous Error Correction in Robotic Grasping Tasks.}
The process begins with the Vision Processing Module, which extracts RGB $\bm{I}$ and depth images $\bm{I_D}$, generating 3D spatial positions $\bm{S}$. $\bm{I}$ pass to the LVLM $\mathcal{M}$, which, along with text instructions $\bm{Ins}$, generates actions $\bm{I_{act}}$ through the Action Module. The Judgment Module evaluates the grasp based on the grasp state $\bm{G_S}$ and grasp position $\bm{G_P}$, using action images $\bm{I_{act}}$,$\bm{S}$, and $\bm{Ins}$. If the grasp fails (i.e., $\bm{G_S} \cup \bm{G_P} = 0$), the system engages the Reflective Reasoning Module, which analyzes the error and proposes corrective suggestions $\bm{R'}$ through reasoning steps. The memory module stores object descriptions and successful grasp strategies to enhance future grasping attempts, ensuring continuous improvement. If the grasp is successful (i.e., $\bm{G_S} \cup \bm{G_P} = 1$), the correct strategy is saved in memory for future reference.
}}
    \label{fig:pipeline}
    \vspace{-1.5em}
\end{figure*}

In this section, we first formulate the problem (Sec. \ref{m1}) and then introduce RoboReflect, a framework for autonomous reflection and correction of robotic actions in the ambiguous grasping scenario. 
The framework comprises four modules: the visual processing and action modules generate robot actions based on visual observations and instructions (Sec. \ref{m0}); the judgment module assesses task completion from environmental feedback (Sec. \ref{m2}); the reflective reasoning module analyzes failures using LVLM (Sec. \ref{m3}); and the memory module records successful strategies to prevent future errors (Sec. \ref{m5}). The framework is shown in Fig. \ref{fig:pipeline}.

\subsection{Problem Statement} \label{m1} 
To support the embodied robot in identifying the objects of ambiguous conditions and facilitating reflective learning, we redefine the grasping problem more precisely.
Unlike traditional grasping tasks that only focus on whether the grasp is completed, we decompose the evaluation of a successful robotic arm grasp event into two components: ``whether the object is successfully grasped'' and ``whether the grasp position is correct'', represented by \bm{$G_S$} and \bm{$G_P$}, respectively. The grasp is considered successful only when both conditions are completed, which can be presented as:
\begin{equation}
\bm{G_{S}} \cup \bm{G_{P}} =
\begin{cases}
1 & \text{if } \bm{G_{S}} = \bm{G_{P}} = 1 \\
0 & \text{if } \bm{G_{S}} = 0 \text{ or } \bm{G_{P}} = 0
\end{cases}
\end{equation}

To obtain a binary array $[\bm{G_S}, \bm{G_P}]$ for decision-making, we need to establish the entire processing pipeline from input to output, as shown in Fig. \ref{fig:pipeline}. 
The entire process has four steps: First, the Visual Processing module ($\bm{VP}$), which converts the input data of RGB-D data to the RGB images and 3D space positions. 
Then, the action module ($\bm{AM}$) transforms input data into actions via the LVLM $\mathcal{M}$;
and following the Judgement module ($\bm{JM}$).
At least, we propose a Reflective Reasoning module ($\bm{RRM}$), which is a novel approach that enables the reflection on grasp results and the correction of erroneous outcomes.

\subsection{Converting RGB-D images \& Action Planning} \label{m0} 
\noindent \textbf{Visual processing module ($\bm{VP}$).} To establish object-centric spatial representations for embodied interaction, our framework hierarchically processes visual observations through three sequential transformations. The process begins with the vision-language model $\mathcal{M}$ analyzing RGBD inputs $\bm{I} \in \mathbb{R}^{H \times W \times 4}$ to detect $N$ objects $\mathcal{O} = \{o_i\}_{i=1}^N$, generating both semantic captions $\bm{C}$ and preliminary 2D bounding boxes $\bm{B}_{\text{2D}} \in \mathbb{R}^{N \times 4}$ through its frozen detection head. These 2D coordinates then serve as geometric prompts for SAM\cite{kirillov2023segment}, which produces pixel-precise instance masks that refine object boundaries beyond the initial detection. By back-projecting the masked depth data via camera intrinsics, we get the 3D position and bounding boxes of each object, collectively referred to as the 3D spatial position $\bm{S}$.

\noindent \textbf{Action module ($\bm{AM}$)} serves as the core component of the robotic execution system, responsible for translating high-level instructions into executable motion sequences for robotic manipulators. In our embodied agent implementation, we systematically encapsulated a set of atomic actions (e.g., ``Move'', ``GraspOn'' and ``GraspOff'') through predefined API libraries, establishing fundamental units for task decomposition and recomposition. The system parses natural language instructions $\bm{Ins}$ by analyzing scene information $[\bm{I},\bm{S}]$. For instance, the command “pick up the water cup” would be decomposed into an atomic action sequence: Move('cup'), GraspOn() and Move().
Notably, during the execution of the action, $\bm{AM}$ records the visual data collected by the RGB camera at the end of the robot arm. These action execution image sequences $\bm{I_{act}}$ provide the system with visual evidence for action verification.

\subsection{Status Determining by Judgment Module} \label{m2}  
After obtaining $\bm{I_{act}}$ as action information, we introduce a Judgment Module ($\bm{JM}$) for determining whether the grasping action is complete and successful.

In the action process, the images $\bm{I_{act}}$ is stored to present the state of the environment action.
In the $\bm{JM}$, the last frame image $\bm{I_{act}}$, along with the instruction text prompt $\bm{Ins}$ and 3D bounding box $\bm{S}$, are encapsulated in a prompt to provide the LVLM $\mathcal{M}$ with the necessary information to infer the task finished status. 
The $\bm{JM}$ then evaluates the grasping success status based on two predefined criteria $[\bm{G_{S}}, \bm{G_{P}} ]$.
When dealing with objects under ambiguous conditions, the judgment primarily hinges on the following aspects:
1) Unsuccessful grasping often stems from changes in the object's properties during manipulation, such as deformations or structural separations, which ultimately lead to failure. 
2) Incorrect grasping positions, while possibly effective for manipulation, are generally considered inappropriate by human experience. This includes grasping edible parts or areas that may pose safety risks, making such positions undesirable from a functional and safety perspective.
Therefore, along with $\bm{I_{act}}$ and $\bm{Ins}$ into the LVLM, two additional textual questions are input into the LVLM module: 1) Was the robotic arm’s grasp successful? 2) Does the grasping position align with human experience?

When the $\bm{JM}$ determines that the task is unsuccessful, \textit{i.e.} $\bm{G_{S}} \cup \bm{G_{P}}=0$, it forwards the action images $\bm{I_{act}}$, $\bm{Ins}$ to requests the Reflective Reasoning Module to initiate reflection (c.f. Sec. \ref{m3}). 
This core component of the framework represents the first step toward autonomous reflection and reasoning.
If the evaluation is deemed successful, \textit{i.e.} $\bm{G_{S}} \cup \bm{G_{P}}=1$, by the $\bm{JM}$, the object $\bm{O_i}$ description information will be stored within the memory module (c.f. Sec. \ref{m5}) for generalizing to grasp other objects in the future task.

\subsection{Correction by Reflective Reasoning Module} \label{m3} 
In this section, we introduce the Reflective Reasoning Module ($\bm{RRM}$), which consists of two sub-modules, the \textit{Self-Reflective module} and the \textit{Discussion module}.

\noindent \textbf{Self-Reflective Module.} When the judgment module determines the grasping failure, analyzing the cause of the failure becomes the key focus which will assist the robot to correct the action.
For objects in ambiguous conditions where the LVLM $\mathcal{M}$ cannot accurately determine their properties, the robotic arm often struggles to execute the correct grasping action. 
In such cases, a lot of key information must be inferred through interaction with the object and observation of feedback in order to perform the correct grasp pose.
Therefore, we introduce a self-reflective module to analyze and discuss the specific causes of grasping failures, wrapping them into prompts understandable by the LVLM $\mathcal{M}$. 
The inputs to the self-reflective module include not only the environment feedback $\bm{I_{act}}$ but also the task's input object's description $\bm{O}$ and text instructions $\bm{Ins}$ as prior information.

Understanding objects under ambiguous conditions and identifying the root cause of errors remains a challenge for LVLMs, even with feedback. To enhance reasoning and improve accuracy, we apply the Chain of Thought (CoT) to assist the robot in introspective reasoning. The reflection result is denoted as $\bm{R}=\{\bm{Y},\bm{P}\}$, where $\bm{Y}$ represents the error cause and $\bm{P}$ suggests corrective actions.

As shown in Fig.~\ref{fig:pipeline}, the LVLM $\mathcal{M}$ first analyzes the object description $O$ to understand objects intrinsic properties.
Next, $\mathcal{M}$ considers the object's context within the environment, linking potential states to the action images $\bm{I_{act}}$.
Given that ambiguous objects often pose two main challenges—changes in object properties and sub-optimal grasping positions—$\mathcal{M}$ is guided to reason in these two directions when determining the error cause $\bm{Y}$.
When the LVLM identifies the cause of the error and constrains it through the intrinsic properties of the object description, the model returns the action corrections description $\bm{P}$. 
At the same time, we combine the identified error description $\bm{Y}$ and $\bm{P}$ to form $\bm{R} = \{ \bm{Y}, \bm{P} \}$, which is then output as the result.
% Once the LVLM $\mathcal{M}$ identifies the error cause and proposes action corrections $\bm{P}$ through reasoning, these suggestions must comply with the inferred constraints of the object. This process yields the final reflection result $\bm{R}$ for the robot.

\begin{figure}[tbp!]
    \centering
    \includegraphics[width=0.48\textwidth]{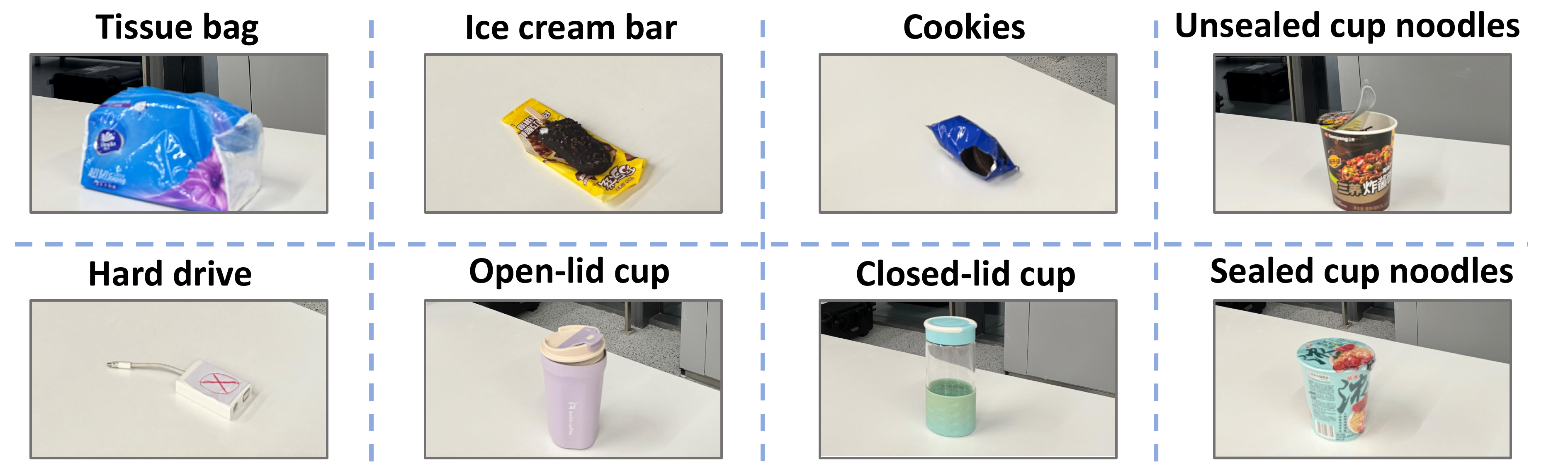}
    \caption{\small{\textbf{Objects descriptions.} The upper half of \textbf{tissue bag} is often empty due to the exhaustion of materials, which makes them susceptible to deformation.  
    A \textbf{closed-lid cup} has a securely fastened lid.
    An \textbf{open-lid cup} risks lid-body separation due to loose closure.
    \textbf{Cookies} are fragile and breakable.
    The upper half of a \textbf{hard drive} is labeled Untouchable.
    \textbf{Sealed cup noodles} denote standard intact packaging.
    \textbf{Unsealed cup noodles} have vulnerable tops prone to spills/deformation.
    The edible portion of an \textbf{ice cream bar} must remain untouched.
    }}
    \label{fig:objs}
    \vspace{-0.5em}
\end{figure}

\noindent \textbf{Discussion Module.} During the Self-Reflective Module, relying on the step-by-step reasoning of the CoT may lead to cumulative errors, which can amplify and propagate throughout the reasoning process \cite{sun2023corex,xu2022learning}.

To enable RoboReflect to robustly handle objects in various ambiguous conditions, we draw inspiration from recent efforts in peer-rating by LLMs\cite{zheng2024judging}, and introduce the Discussion Module with another LVLM $\mathcal{M}_{D}$ to supervise the results of the Self-Reflective Module.
More specifically, after completing the introspective reasoning process, instead of immediately planning new actions, we utilize a module to evaluate the result $R$.
This discussion is formalized as $\bm{R^{\prime}} = \mathcal{M}_{D}(\bm{R}, \bm{I_{act}, \bm{Ins}})$, where $\bm{R^{\prime}}$ encapsulates the discussion outcome.

The discussion process primarily involves two steps,, as shown in the Figure~\ref{fig:pipeline}:
The first step is to evaluate the correctness of $\bm{R}$ by the discussion LVLM $\mathcal{M}_{D}$ with multi-turns Q\&A. 
If $\bm{R}$ is determined to be correct by the LVLM $\mathcal{M}$, the reflection result will be retained. 
However, if any issues are identified, we will proceed to update and improve $\bm{R}$.
Essentially, this approach is premised on a re-analysis of the causes of errors based on the results of reflection and reasoning. It serves to enhance our understanding of objects under ambiguous conditions, thereby avoiding any oversight of grasping constraints.
The reflect suggestion $\bm{R^{\prime}}$, is fed back to the LVLM $\mathcal{M}$ for next time grasp trying. 
It serves as an explanation for the errors encountered and an understanding of objects under ambiguous conditions, thereby assisting the model in planning accurate grasping actions.

\subsection{Memory Module} \label{m5} Successful task execution signifies that the system has mastered the operable attributes of the object. To avoid wasteful repetition of reflection and to facilitate efficient and accurate task completion in the future, it is crucial to retain the results of reflection for subsequent use.
Therefore, we employ a memory module to record the details of successfully executed tasks. This will serve as action experience before assisting future task execution. 

The memory is structured as the `dictionary' data structure, where each `key' represents an object's description $\bm{O}$ with which there has been interaction, and the corresponding `value' encapsulates the understanding $R$ derived from interacting with the object. 
Given that the same object may present different ambiguous conditions in different scenarios, to prevent memory deception leading to errors, each memory is only associated with the actions of that stage. If the scenario changes, all memories are cleared.

%% file: sections/experiments.tex
\section{EXPERIMENTS}

\begin{figure}[t!]
    \centering
    \includegraphics[width=0.48\textwidth]{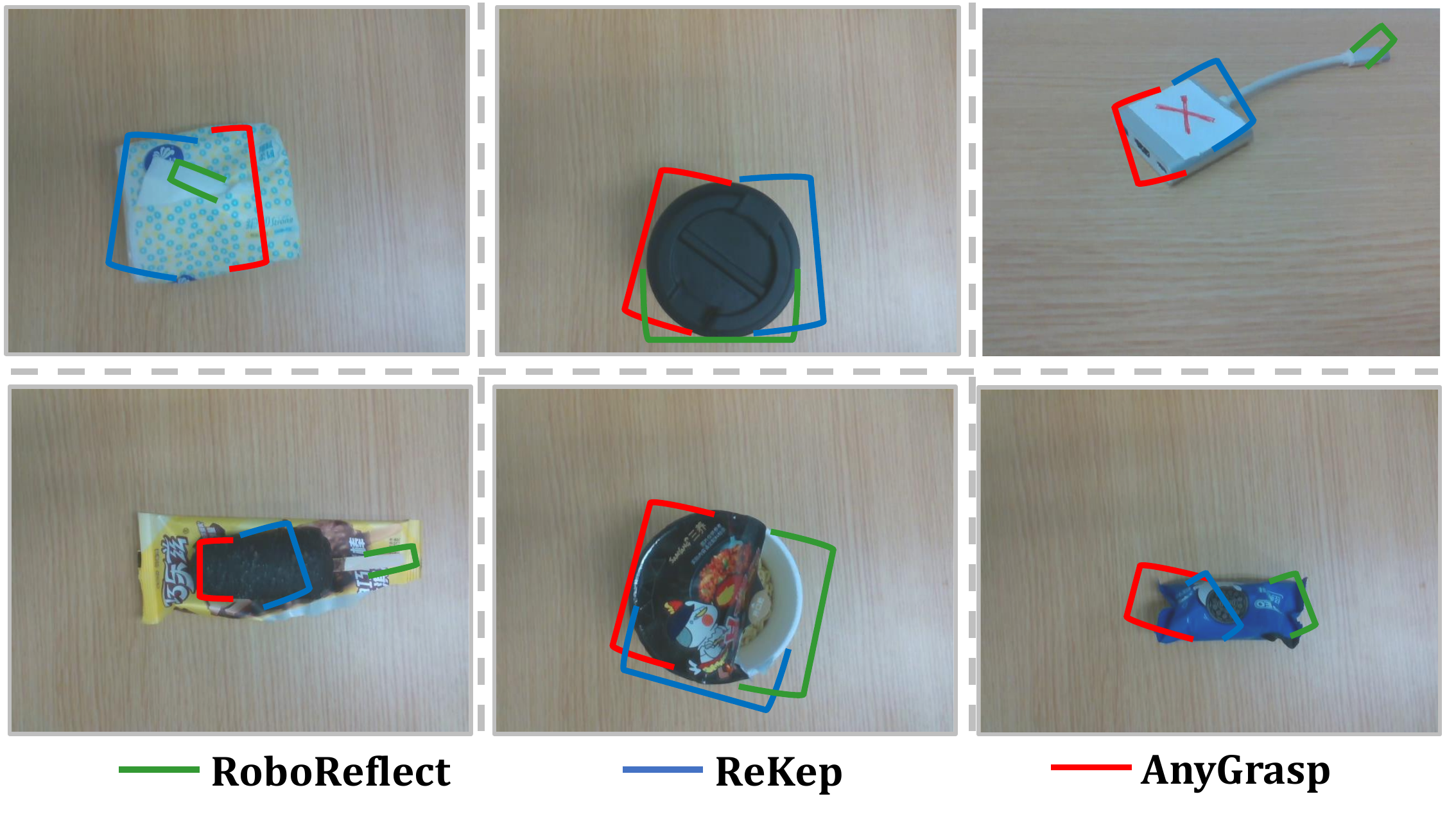}
    \caption{\small{\textbf{Visual comparison of grasping posture.} The green ones are our method, the blue poses are the result of ReKep, and the red ones are the result of AnyGrasp. }}
    \label{fig:res}
    \vspace{-2em}
\end{figure}

\begin{figure*}[t!]
    \centering
    \includegraphics[width=\textwidth]{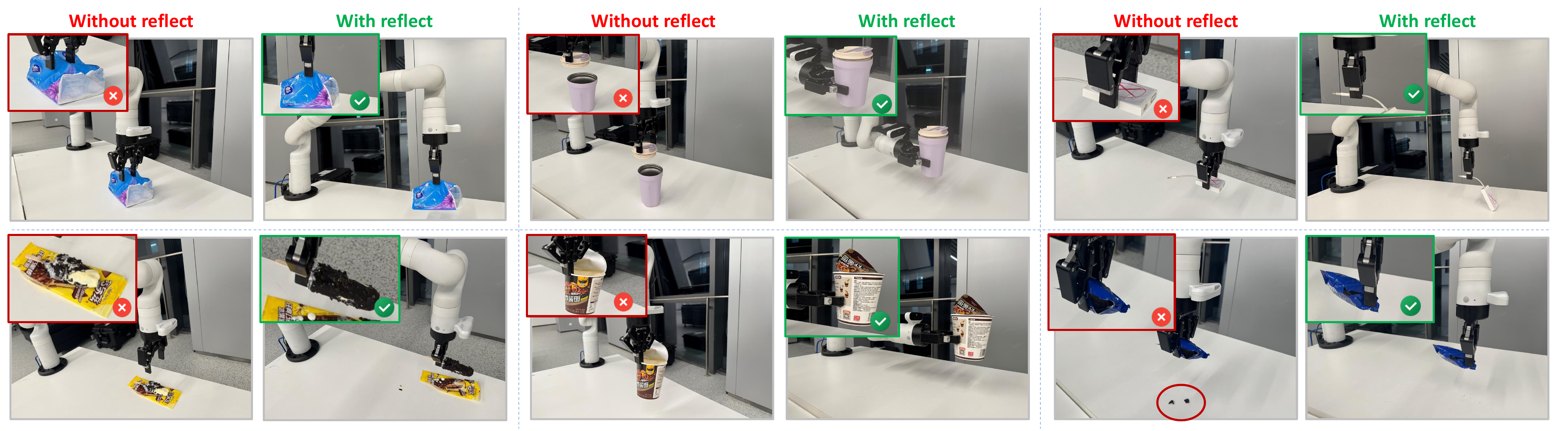}
    \caption{\small{\textbf{The comparison between failed (without reflect) and successful (with reflect) grasping cases.} We present the grasping results of six objects before and after using our RoboReflect model. Before reflection, the grasping attempts often failed due to an inability to understand the object's properties, resulting in either a failure to grasp or an incorrect grasp position. After reflection, the model was able to successfully grasp the objects.}}
    \label{fig:qual}
    \vspace{-0.5em}
\end{figure*}

\begin{table*}[t]
    \caption{\small{Performance of RoboReflect and baselines on eight ambiguous-condition objects. Note that the numbers in parentheses following the RoboReflect results indicate in which attempt the error occurred.}}
    \centering
    \resizebox{1\textwidth}{!}{%
    \begin{tabular}{c| c c c c c c c c} 
        \toprule 
        Method & Tissue bag & Ice cream bar & Cookies & Sealed cup noodles & Unsealed cup noodles & Closed-lid cup & Open-lid cup & Hard drive \\
        
        \midrule 
        GPT-4V \cite{achiam2023gpt} & 0$\%$ & 0$\%$ & 10$\%$ & 60$\%$ & 0$\%$ & 70$\%$ & 10$\%$ & 0$\%$ \\
        AnyGrasp \cite{fang2023anygrasp}  & 20$\%$ & 30$\%$ & 20$\%$ & 80$\%$ & 70$\%$ & 80$\%$ & 80$\%$ & 30$\%$ \\
        ReKep \cite{huang2024rekep}  & 10$\%$ & 50$\%$ & 10$\%$ & \textbf{90$\%$} & \textbf{80$\%$} & \textbf{80$\%$} & \textbf{90$\%$} & 10$\%$ \\
        \midrule
        \textbf{RoboReflect} & \textbf{70$\%$} (1,2,4) & \textbf{60$\%$} (1,4,7,8) & \textbf{60$\%$} (1,2,4,5) & \textbf{90$\%$} (1) & \textbf{80$\%$} (1,2) & \textbf{80$\%$} (2,6) & 70$\%$ (1,2,3) & \textbf{60$\%$} (1,2,5,9) \\
        \bottomrule
    \end{tabular}
    }
    \label{tab:overall}
    \vspace{-1.5em}
\end{table*}

\subsection{Experimental Setup}
\noindent \textbf{Hardware.} We use a Kinova Gen3 Ultra lightweight robot with a 7-DoF arm and a Robotiq 2F-140 gripper in our experiments. 
For image and depth perception, we apply a RealSense D400 depth camera mounted on a wrist to capture RGB-D images.
Since the experiment involves grasping only one or two objects, there are no occlusion issues. Therefore, we did not include additional camera setups.

\noindent \textbf{Tasks Setting.}
As shown in Fig.~\ref{fig:objs}, we select eight common objects from daily life to represent objects with ambiguous conditions. Since the experiment focuses solely on grasping tasks, and to simulate the ambiguously associated with grasping unfamiliar objects, we standardized the initial grasping pose as approaching the object from directly above.

\noindent \textbf{LVLM.} We use GPT-4V from OpenAI API as our LVLM.
To ensure the comparison is fair, apart from processing inputs from other modules during the reflection process, we do not include any in-context examples in the prompt.

\noindent \textbf{Metrics.} To quantify our performance, we use the success rate of grasping ambiguous-condition objects for evaluation.
The success rate not only considers the grasping state $\bm{G_S}$ but also the grasping position $\bm{G_P}$ correct or not.
The correct grasping position $\bm{G_P}$ of objects is always defined by the human experience. 
Based on the inherent human experience of the ambiguous condition grasping, we categorize objects prone to ambiguous conditions into three types, the details descriptions are shown in Fig.~\ref{fig:objs}:
\begin{itemize}
    \item \textbf{Soft objects with deformable surfaces}: When the object's functional properties change, the grasping position and force need to be reconsidered. For example, when half of a tissue bag has been used, the grasping strategy must differ from that of an unopened tissue bag. Similarly, the approach for grasping an unopened cup of instant noodles differs from one that has been opened and filled with hot water.
    
    \item \textbf{Assembled objects composed of different parts}: Grasping different parts of the object can lead to varying outcomes. For instance, an open-lid cup cannot be grasped entirely if only the lid is grabbed from above. Similar cases include teapots with lids, etc.
    
    \item \textbf{Objects with inherent properties that prevent certain parts from being grasped}: Grasping these parts would render the object unusable. For example, an ice cream bar should only be grasped by its stick, while a hardware drive should be handled by its lower, less fragile part to avoid damaging the main component.
\end{itemize}

\subsection{Overall Performance}
To demonstrate the advantages of our framework, we compare it with ReKep \cite{huang2024rekep}, AnyGrasp \cite{fang2023anygrasp} as baselines and a high-level policy planner using GPT-4V \cite{achiam2023gpt}, which excludes other framework components.
We evaluate the performance of RoboReflect against baseline methods and compare their grasping success rates on the eight objects, as shown in Table \ref{tab:overall}. 
The results indicate that both ReKep, AnyGrasp and GPT-4V performed poorly when dealing with ambiguous condition objects, with low success rates, demonstrating that this category of objects presents a challenge for general grasping algorithms. RoboReflect shows a significant advantage compared to the baseline methods. Compared to the AnyGrasp, RoboReflect improves the average accuracy by 21.25\%; compared to GPT-4V, it improves by 50\% and compared to ReKep it improves by 17.5\%. These results highlight the exceptional performance of RoboReflect when grasping ambiguous conditions objects.
\begin{table*}[!ht]
    \caption{\small{Ablation study results for RoboReflect on eight ambiguous-condition objects.}}
    \centering
    \resizebox{1\textwidth}{!}{%
    \begin{tabular}{c|c c c c c c c c} 
        \toprule  
        Method & Tissue bag & Ice cream bar & Cookies & Sealed cup noodles & Unsealed cup noodles & Closed-lid cup & Open-lid cup & Hard drive \\
        \midrule 
        w/o Discuss. & 60$\%$ & 60$\%$ & 40$\%$ & 80$\%$ & 60$\%$ & 80$\%$ & 50$\%$ & 20$\%$ \\
        %\midrule
        \textbf{RoboReflect} & 70$\%$ & 60$\%$ & 60$\%$ & 90$\%$ & 80$\%$ & 80$\%$ & 70$\%$ & 60$\%$ \\
        \bottomrule
    \end{tabular}
    }
    \label{tab:wo_dis}
    \vspace{-2em}
\end{table*}

\begin{table}[!ht]
    \caption{\small{Ablation study results for RoboReflect, each class contains two objects with different states.} }
    \centering
    \resizebox{0.43\textwidth}{!}{
    \begin{tabular}{*{3}{c}} 
        \toprule  Method & Mixed Cup & Mixed Cup noodles \\
        \midrule 
        RoboReflect w/o Mem. & 75$\%$ & 80$\%$ \\
        \textbf{RoboReflect w/ Mem.} & 90$\%$ & 95$\%$ \\
        %\midrule
        \bottomrule
    \end{tabular}}
    \label{tab:wo_mem}
    \vspace{-2.5em}
\end{table}
When grasping a partially empty tissue bag, the top half is prone to deformation, causing grasping failures regardless of how AnyGrasp or GPT-4V attempts to handle it.
However, the extremely high success rate (70\%) of RoboReflect highlights the superiority of our framework. 
When grasping cookies, external hard drives and ice cream bars present greater challenges to the baseline methods. 
The grasping of these objects contains some ambiguous conditions, which may be common sense to humans, but difficult to understand for baseline methods. 
Although ReKep demonstrates the capability to handle basic object manipulation tasks, such as grasping ice cream, its accuracy of 50\% remains slightly inferior to that of our proposed method, its grasping performance is significantly limited when faced with objects that require more understanding, such as cookies and external hard drives, with a success rate of only 10\%.
Therefore, the success rate of AnyGrasp and GPT-4V is extremely low, as 10\%, 0\%, 0\% and 20\%, 30\%, 30\%, respectively. 
In contrast, RoboReflect can overcome these ambiguous conditions through autonomously reflective, thus achieving correct grasping with high performance, 60\%, on three objects.

It's worth noting that when grasping cup noodles and closed-lid cups, the performance of RoboReflect is not much different from that of ReKep, AnyGrasp and GPT-4V, with similar success rates. This is because these two objects are in an ideal state and not prone to characteristic changes, so there is no need to think too much about the grasping approach. 
However, when grasping open-lid cups and open-lid cup noodles, the success rate of ReKep and AnyGrasp has a similar performance to RoboReflect, but much higher than GPT-4V. 
This is because AnyGrasp initially tends to grasp cup-like objects from the body of the cup, while our method initially grasps from directly above and needs to reflect and reason to master the correct grasping method.
To present the experimental results more intuitively, Fig.~\ref{fig:res} shows the comparison of the grasping pose generation effects of AnyGrasp, ReKep and our method on these objects.
We also show some qualitative results of RoboReflect in Fig.~\ref{fig:qual}.

\subsection{Reflect Analysis}
We have compiled statistics on the number of attempts it took to fail when grasping each object. The results, as shown in Table \ref{tab:overall}, provide an intuitive understanding of the effectiveness of the RoboReflect. We observe that due to the initial grasping pose settings, the first attempt to grasp each object failed. However, most errors occurred in the first few attempts, This is attributed to RoboReflect's autonomously reflective capability. Further analysis revealed that these objects can be roughly divided into two categories: easy-to-reflect and hard-to-reflect:
1) Easy-to-reflect objects: tissue bags, sealed cup noodles, unsealed cup noodles, open-lid cups, closed-lid cups, and ice cream bars. The reasons for errors with these objects are obvious, and usually, only one or two autonomous reflective steps to successfully complete the task. 
2) Hard-to-reflect objects: The remaining three objects require three to four reflection steps due to more complex ambiguous conditions. For instance, cookie packaging is difficult to recognize from certain angles, requiring multiple failures and reflections for successful grasping. Similarly, the hard drive, despite having clearly marked non-touchable areas, remains challenging for RoboReflect due to its small graspable area, contributing to the high failure rate.
In general, each object has its own characteristics, and it is challenging to complete the grasping task in one go. However, through autonomously reflective, RoboReflect can ultimately achieve tasks with an impressive success rate.

\subsection{Ablation Study}
We conduct ablation studies to validate the effectiveness of the Discussion module and Memory module in the pipeline.

\noindent \textbf{Function of Discussion module.} Without our discussion module, the success rate of grasping eight objects is as shown in Table, with the average success rate decreasing by 15.2\%. It can be seen that the success rates on hard-to-reflect objects, such as cookies and external hard drives, have decreased the most, dropping by 20\%. 
This improvement demonstrates that our discussion module enables the LVLM to reason through complex ambiguous object conditions, enhancing understanding and accelerating problem identification.

\noindent \textbf{Importance of Memory module.} 
To evaluate the key role of the Memory Module in our model, we designed the following experiment: 
First, through the reflective reasoning module (c.f. Sec. \ref{m3}), we successfully grasp the corresponding objects. Next, without removing the memory, we attempt 20 mixed grasps of open-lid and closed-lid cups, calculating the success rate.
Then, we remove the memory module and repeat the 20 mixed grasps of open-lid and closed-lid cups, calculating the success rate again.
The same mixed experiment is conducted on cup noodles, with and without the memory module, grasping both sealed and unsealed cup noodles. The results of the experiment are shown in Table~\ref{tab:wo_mem}.
When grasping with memory, the success rate for grasping both types of objects exceeded 90\%, reaching a usable level.
However, after removing the memory module, the absence of stored prior information in memory makes it more difficult for the LVLM to recognize the state of the cup or whether the cup noodle is sealed, leading to less effective grasping strategies and a drop in accuracy. This experiment clearly demonstrates the importance of the Memory Module in RoboReflect for maintaining high grasping accuracy.

\section{CONCLUSION}
In this paper, we introduced RoboReflect, a novel framework for grasping robots that enables autonomous reflection and action correction using large vision-language models (LVLMs). By addressing the challenges posed by ambiguous-condition objects, RoboReflect enhances robotic grasping performance through continuous reflection and strategy refinement. Our experiments on eight common objects across three categories demonstrate that RoboReflect significantly improves grasp success rates compared to existing methods. The framework's capability to store successful strategies for future reference further strengthens its capability to handle diverse scenarios without human intervention. These results highlight the importance of incorporating autonomous reflection into robotic systems, offering a promising solution for more resilient and adaptable robots in real-world applications.

%% file: root.bbl
\begin{thebibliography}{10}
\providecommand{\url}[1]{#1}
\csname url@rmstyle\endcsname
\providecommand{\newblock}{\relax}
\providecommand{\bibinfo}[2]{#2}
\providecommand\BIBentrySTDinterwordspacing{\spaceskip=0pt\relax}
\providecommand\BIBentryALTinterwordstretchfactor{4}
\providecommand\BIBentryALTinterwordspacing{\spaceskip=\fontdimen2\font plus
\BIBentryALTinterwordstretchfactor\fontdimen3\font minus \fontdimen4\font\relax}
\providecommand\BIBforeignlanguage[2]{{%
\expandafter\ifx\csname l@#1\endcsname\relax
\typeout{** WARNING: IEEEtran.bst: No hyphenation pattern has been}%
\typeout{** loaded for the language `#1'. Using the pattern for}%
\typeout{** the default language instead.}%
\else
\language=\csname l@#1\endcsname
\fi
#2}}

\bibitem{wang2024llm}
R.~Wang, Z.~Yang, Z.~Zhao, X.~Tong, Z.~Hong, and K.~Qian, ``Llm-based robot task planning with exceptional handling for general purpose service robots,'' \emph{arXiv preprint arXiv:2405.15646}, 2024.

\bibitem{guan2024atom}
I.~Y. Guan, G.~Zhang, X.~Liu, E.~Zhao, and J.~Wu, ``Atom: Leveraging large language models for adaptive task object motion strategies in object rearrangement for service robotics,'' in \emph{2024 10th International Conference on Electrical Engineering, Control and Robotics (EECR)}.\hskip 1em plus 0.5em minus 0.4em\relax IEEE, 2024, pp. 8--13.

\bibitem{macaluso2024toward}
A.~Macaluso, N.~Cote, and S.~Chitta, ``Toward automated programming for robotic assembly using chatgpt,'' \emph{arXiv preprint arXiv:2405.08216}, 2024.

\bibitem{liu2023reflect}
Z.~Liu, A.~Bahety, and S.~Song, ``Reflect: Summarizing robot experiences for failure explanation and correction,'' \emph{arXiv preprint arXiv:2306.15724}, 2023.

\bibitem{shi2024yell}
L.~X. Shi, Z.~Hu, T.~Z. Zhao, A.~Sharma, K.~Pertsch, J.~Luo, S.~Levine, and C.~Finn, ``Yell at your robot: Improving on-the-fly from language corrections,'' \emph{arXiv preprint arXiv:2403.12910}, 2024.

\bibitem{huang2024rekep}
W.~Huang, C.~Wang, Y.~Li, R.~Zhang, and L.~Fei-Fei, ``Rekep: Spatio-temporal reasoning of relational keypoint constraints for robotic manipulation,'' \emph{arXiv preprint arXiv:2409.01652}, 2024.

\bibitem{fang2023anygrasp}
H.-S. Fang, C.~Wang, H.~Fang, M.~Gou, J.~Liu, H.~Yan, W.~Liu, Y.~Xie, and C.~Lu, ``Anygrasp: Robust and efficient grasp perception in spatial and temporal domains,'' \emph{IEEE Transactions on Robotics}, 2023.

\bibitem{achiam2023gpt}
J.~Achiam, S.~Adler, S.~Agarwal, L.~Ahmad, I.~Akkaya, F.~L. Aleman, D.~Almeida, J.~Altenschmidt, S.~Altman, S.~Anadkat, \emph{et~al.}, ``Gpt-4 technical report,'' \emph{arXiv preprint arXiv:2303.08774}, 2023.

\bibitem{chowdhery2023palm}
A.~Chowdhery, S.~Narang, J.~Devlin, M.~Bosma, G.~Mishra, A.~Roberts, P.~Barham, H.~W. Chung, C.~Sutton, S.~Gehrmann, \emph{et~al.}, ``Palm: Scaling language modeling with pathways,'' \emph{Journal of Machine Learning Research}, vol.~24, no. 240, pp. 1--113, 2023.

\bibitem{touvron2023llama}
H.~Touvron, T.~Lavril, G.~Izacard, X.~Martinet, M.-A. Lachaux, T.~Lacroix, B.~Rozi{\`e}re, N.~Goyal, E.~Hambro, F.~Azhar, \emph{et~al.}, ``Llama: Open and efficient foundation language models,'' \emph{arXiv preprint arXiv:2302.13971}, 2023.

\bibitem{ahn2022can}
M.~Ahn, A.~Brohan, N.~Brown, Y.~Chebotar, O.~Cortes, B.~David, C.~Finn, C.~Fu, K.~Gopalakrishnan, K.~Hausman, \emph{et~al.}, ``Do as i can, not as i say: Grounding language in robotic affordances,'' \emph{arXiv preprint arXiv:2204.01691}, 2022.

\bibitem{lin2023text2motion}
K.~Lin, C.~Agia, T.~Migimatsu, M.~Pavone, and J.~Bohg, ``Text2motion: From natural language instructions to feasible plans,'' \emph{Autonomous Robots}, vol.~47, no.~8, pp. 1345--1365, 2023.

\bibitem{singh2023progprompt}
I.~Singh, V.~Blukis, A.~Mousavian, A.~Goyal, D.~Xu, J.~Tremblay, D.~Fox, J.~Thomason, and A.~Garg, ``Progprompt: Generating situated robot task plans using large language models,'' in \emph{2023 IEEE International Conference on Robotics and Automation (ICRA)}.\hskip 1em plus 0.5em minus 0.4em\relax IEEE, 2023, pp. 11\,523--11\,530.

\bibitem{liang2023code}
J.~Liang, W.~Huang, F.~Xia, P.~Xu, K.~Hausman, B.~Ichter, P.~Florence, and A.~Zeng, ``Code as policies: Language model programs for embodied control,'' in \emph{2023 IEEE International Conference on Robotics and Automation (ICRA)}.\hskip 1em plus 0.5em minus 0.4em\relax IEEE, 2023, pp. 9493--9500.

\bibitem{kim2024openvla}
M.~J. Kim, K.~Pertsch, S.~Karamcheti, T.~Xiao, A.~Balakrishna, S.~Nair, R.~Rafailov, E.~Foster, G.~Lam, P.~Sanketi, \emph{et~al.}, ``Openvla: An open-source vision-language-action model,'' \emph{arXiv preprint arXiv:2406.09246}, 2024.

\bibitem{zhen20243d}
H.~Zhen, X.~Qiu, P.~Chen, J.~Yang, X.~Yan, Y.~Du, Y.~Hong, and C.~Gan, ``3d-vla: A 3d vision-language-action generative world model,'' \emph{arXiv preprint arXiv:2403.09631}, 2024.

\bibitem{li2023vision}
X.~Li, M.~Liu, H.~Zhang, C.~Yu, J.~Xu, H.~Wu, C.~Cheang, Y.~Jing, W.~Zhang, H.~Liu, \emph{et~al.}, ``Vision-language foundation models as effective robot imitators,'' \emph{arXiv preprint arXiv:2311.01378}, 2023.

\bibitem{shinn2303reflexion}
N.~Shinn, B.~Labash, and A.~Gopinath, ``Reflexion: an autonomous agent with dynamic memory and self-reflection. arxiv (2023) doi: 10.48550,'' \emph{arXiv preprint arxiv.2303.11366}, 2023.

\bibitem{huang2022inner}
W.~Huang, F.~Xia, T.~Xiao, H.~Chan, J.~Liang, P.~Florence, A.~Zeng, J.~Tompson, I.~Mordatch, Y.~Chebotar, \emph{et~al.}, ``Inner monologue: Embodied reasoning through planning with language models,'' \emph{arXiv preprint arXiv:2207.05608}, 2022.

\bibitem{ren2023robots}
A.~Z. Ren, A.~Dixit, A.~Bodrova, S.~Singh, S.~Tu, N.~Brown, P.~Xu, L.~Takayama, F.~Xia, J.~Varley, \emph{et~al.}, ``Robots that ask for help: Uncertainty alignment for large language model planners,'' \emph{arXiv preprint arXiv:2307.01928}, 2023.

\bibitem{kirillov2023segment}
A.~Kirillov, E.~Mintun, N.~Ravi, H.~Mao, C.~Rolland, L.~Gustafson, T.~Xiao, S.~Whitehead, A.~C. Berg, W.-Y. Lo, \emph{et~al.}, ``Segment anything,'' in \emph{Proceedings of the IEEE/CVF International Conference on Computer Vision}, 2023, pp. 4015--4026.

\bibitem{sun2023corex}
Q.~Sun, Z.~Yin, X.~Li, Z.~Wu, X.~Qiu, and L.~Kong, ``Corex: Pushing the boundaries of complex reasoning through multi-model collaboration,'' \emph{arXiv preprint arXiv:2310.00280}, 2023.

\bibitem{xu2022learning}
J.~Xu, X.~Liu, J.~Yan, D.~Cai, H.~Li, and J.~Li, ``Learning to break the loop: Analyzing and mitigating repetitions for neural text generation,'' \emph{Advances in Neural Information Processing Systems}, vol.~35, pp. 3082--3095, 2022.

\bibitem{zheng2024judging}
L.~Zheng, W.-L. Chiang, Y.~Sheng, S.~Zhuang, Z.~Wu, Y.~Zhuang, Z.~Lin, Z.~Li, D.~Li, E.~Xing, \emph{et~al.}, ``Judging llm-as-a-judge with mt-bench and chatbot arena,'' \emph{Advances in Neural Information Processing Systems}, vol.~36, 2024.

\end{thebibliography}
